\documentclass[11pt]{article}

\usepackage[final]{acl}

\usepackage{times}
\usepackage{latexsym}
\usepackage{booktabs}   
\usepackage{multirow}   
\usepackage{graphicx}   
\usepackage{array}      
\usepackage{graphicx}      
\usepackage{subcaption}    
\usepackage{caption}       
\usepackage{float}         
\usepackage{amsmath}
\usepackage{amssymb}   

\usepackage[T1]{fontenc}

\usepackage[utf8]{inputenc}

\usepackage{microtype}

\usepackage{inconsolata}

\usepackage{graphicx}

\title{Iterative Structured Pruning for Large Language Models with Multi-Domain Calibration}

\author{
\bf Guangxin Wu$^{1,2}$\thanks{These authors contribute equally to this work.},
Hao Zhang$^{1,2,3}$\footnotemark[1],
Zhibin Zhang$^{1}$,
Jiafeng Guo$^{1}$,
Xueqi Cheng$^{1}$ \\
\normalsize{$^{1}$Institute of Computing Technology, Chinese Academy of Sciences} \\
\normalsize{$^{2}$University of Chinese Academy of Sciences} \\
\normalsize{$^{3}$School of Advanced Interdisciplinary Sciences, University of Chinese Academy of Sciences} \\
\normalsize{\texttt{(wuguangxin24, zhanghao233)@mails.ucas.ac.cn}} \\
\normalsize{\texttt{(zhangzhibin, guojiafeng, cxq)@ict.ac.cn}}
}

\begin{document}
\maketitle
\begin{abstract}

Large Language Models (LLMs) have achieved remarkable success across a wide spectrum of natural language processing tasks. However, their ever-growing scale introduces significant barriers to real-world deployment, including substantial computational overhead, memory footprint, and inference latency. While model pruning presents a viable solution to these challenges, existing unstructured pruning techniques often yield irregular sparsity patterns that necessitate specialized hardware or software support. In this work, we explore structured pruning, which eliminates entire architectural components and maintains compatibility with standard hardware accelerators. We introduce a novel structured pruning framework that leverages a hybrid multi-domain calibration set and an iterative calibration strategy to effectively identify and remove redundant channels. Extensive experiments on various models across diverse downstream tasks show that our approach achieves significant compression with minimal performance degradation.

\end{abstract}

\section{Introduction}

Large Language Models (LLMs) have demonstrated remarkable capabilities in natural language processing, enabling a wide range of applications such as question answering, summarization, and code generation \cite{ding2022, qin2023,zhu2023,li2023}. Moreover, these models also demonstrate exceptional performance across a wide range of other domains, including medicine \cite{qi2025mediaug,luo2025pathohr,cong2025hierarchical,qi2025medconv}, security \cite{
ma2025cad,wu2025sugar}, and various social tasks \cite{zhang2025can,zhang2025asymoe,zheng2025g2rammar,zheng2025graphgeo}. As model sizes continue to grow, LLMs exhibit emergent behaviors and enhanced reasoning abilities. However, the increasing scale and complexity of these models pose significant challenges for practical deployment. The substantial computational and memory requirements lead to high inference latency, elevated energy consumption, and strict hardware constraints, which limit their usability in resource-constrained or real-time settings \cite{zhang2023,huang2023,wang2023}. These challenges highlight the urgent need for effective model compression and acceleration techniques that align with the unique characteristics of LLMs.

Among various solutions, model pruning \cite{ma2023llm,ashkboos2024slicegpt,li2023communication,han2015deep} has emerged as a particularly promising direction. It can be broadly categorized into unstructured pruning and structured pruning. Unstructured pruning \cite{liao2023unstructured,unstructured_pruning_speech} removes individual weights from parameter matrices, but often results in irregular sparsity patterns that demand specialized hardware and software for efficient execution. This irregularity not only complicates storage and inference but also reduces portability and scalability. Common unstructured approaches evaluate the significance of individual parameters and eliminate those with minimal impact, followed by adjustments to the remaining weights. While effective in some cases, these methods disrupt the model’s structural coherence.

Structured pruning \cite{ashkboos2024slicegpt,yang2022comparative} offers an alternative that addresses these limitations by removing entire architectural components such as neurons, channels, or layers. This type of pruning simplifies the model at a coarser granularity, making the resulting models more compatible with general-purpose hardware and standard deep learning frameworks. It reduces both computational overhead and memory usage while preserving the high-level structure of the original model.

In this work, we present a new structured pruning framework that integrates a hybrid calibration set drawn from multiple domains with an iterative calibration strategy. This design enables accurate identification of redundant channels with minimal loss in model performance. By combining diverse data representations with a progressive pruning process, our method achieves efficient model compression and strong generalization across downstream tasks. Extensive experiments on a variety of LLM architectures demonstrate that our approach outperforms existing structured pruning baselines in terms of both compression ratio and accuracy preservation. Our contributions are summarized as follows:

\begin{itemize}
    \item \textbf{Multi-domain hybrid calibration set.} We design a diverse calibration dataset that spans multiple domains, including Wikipedia articles, Common Crawl data, code repositories, and mathematical texts. This diversity enables the pruning process to generalize more effectively across a wide range of linguistic and semantic patterns.
    
    \item \textbf{Iterative channel selection.} We propose an iterative calibration strategy that incrementally refines the choice of channels to prune. This progressive refinement improves both the accuracy of channel selection and the robustness of the pruned model.
    
    \item \textbf{Comprehensive evaluation.} We evaluate our approach on the Qwen2.5 families using a broad set of downstream tasks and datasets. Our method consistently achieves strong performance while delivering substantial model compression.
\end{itemize}
\begin{figure*}[!htbp]
\centering 
\includegraphics[width=0.7\linewidth]{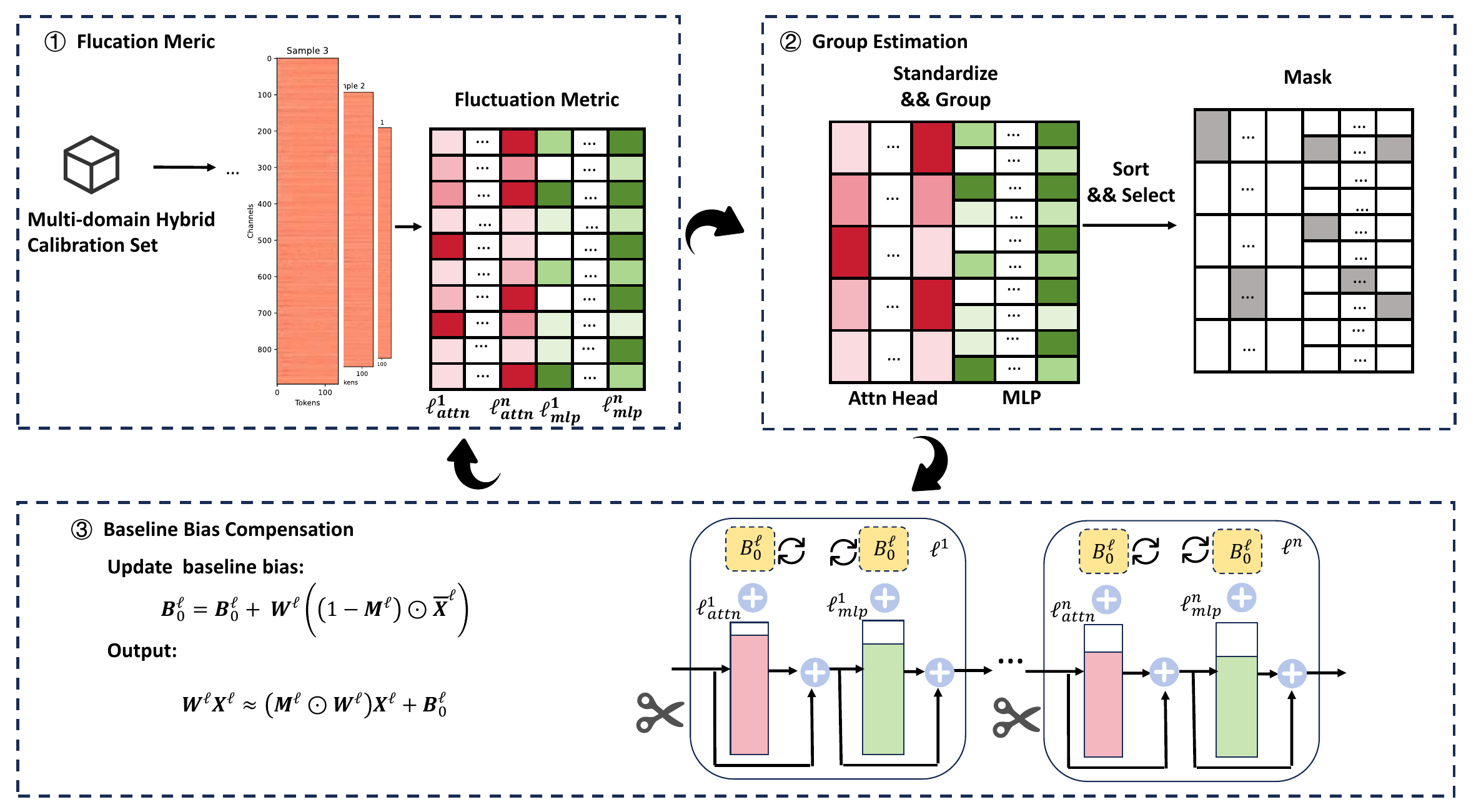}
\caption{Overview of our proposed method.}
\label{1}
\end{figure*}

\section{Related Work}

\subsection{Compression Techniques for Large Language Models}

With the rapid growth of large language models (LLMs) containing billions of parameters, efficient and scalable compression has become increasingly essential. Knowledge distillation \cite{yang2021knowledge,2024CDL}, though effective, is often impractical at this scale due to the high cost of training student models. Quantization methods \cite{zhou2021smoothquant,cai2023gptq,zhou2024framequant} reduce memory and computation by lowering numerical precision, but face challenges in LLMs such as activation outliers and sensitivity to precision errors that can significantly degrade performance.

\subsection{Structured Pruning for Neural Networks}

Network pruning is a long-standing approach for compressing neural networks by removing redundant parameters \cite{ma2023llm,ashkboos2024slicegpt,li2023communication,han2015deep,yang2022comparative}. Early unstructured pruning methods eliminate individual weights based on magnitude or sensitivity, achieving high sparsity but poor hardware efficiency. In contrast, structured pruning removes entire channels, neurons, or attention heads, preserving layer regularity and enabling efficient parallel computation and memory access. Recent advances \cite{ma2023llm} extend structured pruning to transformer architectures, employing criteria such as $\ell_1$ norms, gradient signals, and second-order approximations. Post-training structured pruning further enables compression without full retraining, though lightweight fine-tuning is often required to recover performance after aggressive pruning.

\section{Methodology}
In this section, we present a structured pruning framework for large language models that integrates a variance-based importance criterion from FLAP \cite{an2024flap}, a domain-diverse calibration dataset to enhance generalization across input distributions, and an iterative calibration strategy that refines pruning decisions by accounting for cumulative pruning effects, improving stability and final performance.

\subsection{Preliminary}

Recent studies introduce bias compensation to mitigate pruning-induced output shifts. In structured pruning, the output of an uncompressed layer can be expressed as follows:  
\begin{equation}
\resizebox{0.43\textwidth}{!}{$
W^{\ell} X^{\ell} =
\underbrace{(M^{\ell} \odot W^{\ell}) X^{\ell}}_{\text{Retained Part}} +
\underbrace{((1 - M^{\ell}) \odot W^{\ell}) X^{\ell}}_{\text{Removed Part}}
$}
\end{equation}
where $W^{\ell}$ and $X^{\ell}$ denote the weights and inputs of the $\ell$-th layer, and $M^{\ell} \in \{0,1\}^{\text{shape}(W^{\ell})}$ is a binary mask indicating the retained structures. The goal is to minimize the influence of the removed part,
$\Delta Y^{\ell} = ((1 - M^{\ell}) \odot W^{\ell}) X^{\ell}$, on the output feature map. To compensate for this error, a bias term can be constructed from the mean input activations over tokens and samples for each channel as follows:  
\begin{equation}
\overline{\mathbf{X}}_{:, j,:}^{\ell} =
\frac{1}{N L} \sum_{n=1}^{N} \sum_{k=1}^{L} \mathbf{X}_{n, j, k}^{\ell}
\end{equation}
After determining the pruning mask $M_{\ell}$, the baseline activations of pruned channels are transformed into a bias vector as follows:  
\begin{equation}
\mathbf{B}_0^{\ell} = \mathbf{W}^{\ell} \big((1 - \mathbf{M}^{\ell}) \odot \overline{\mathbf{X}}^{\ell}\big)
\end{equation}
\begin{equation}
\mathbf{W}^{\ell} \mathbf{X}^{\ell} \approx
(\mathbf{M}^{\ell} \odot \mathbf{W}^{\ell}) \mathbf{X}^{\ell} + \mathbf{B}_0^{\ell}
\end{equation}
where $\mathbf{B}_0^{\ell} \in \mathbb{R}^{C_{\text{out}}}$ approximates the output of the original layer. Channel importance depends on both input variance and weight magnitude. A fluctuation metric is defined as follows:  
\begin{equation}
\mathbf{S}_{:, j}^{\ell} =
\frac{1}{N - 1} \sum_{n=1}^{N}
(\mathbf{X}_{n, j,:}^{\ell} - \overline{\mathbf{X}}_{:, j,:}^{\ell})^2
\cdot \|\mathbf{W}_{:, j}^{\ell}\|^2
\end{equation}
and channels with lower fluctuation scores are pruned, with the resulting error compensated by $\mathbf{B}_0^{\ell}$.

Compared to incremental pruning methods that analytically adjust weights after each removal step, this bias-based strategy prunes all target structures in one shot and compensates the output shift using the estimated bias term. It eliminates retraining and is computationally efficient, but its effectiveness depends on accurate activation statistics obtained from calibration data. To enhance robustness, we propose two extensions: (i) constructing a domain-diverse calibration dataset to better capture activation statistics, and (ii) introducing an iterative calibration strategy to mitigate cascading errors in one-shot pruning. These components are detailed below, and Figure~\ref{1} provides an overview of the method.

\subsection{Multi-domain Hybrid Calibration Set}
To enable structured pruning that generalizes across diverse real-world applications, we construct a domain-diverse calibration dataset. Prior pruning methods typically rely on calibration sets from a single or narrow domain, which biases importance estimation toward domain-specific features and reduces robustness in heterogeneous environments where input distributions vary widely.

Formally, consider $K$ distinct domains $\mathcal{D} = \{ \mathcal{D}_1, \ldots, \mathcal{D}_K \}$, each with input distribution $P_k(\mathbf{X})$. For the $\ell$-th layer, the mean activation and variance in domain $k$ are defined as follows:  
\begin{equation}
\resizebox{0.43\textwidth}{!}{$
\overline{\mathbf{X}}^{\ell}_k = \mathbb{E}_{\mathbf{X} \sim P_k}[\mathbf{X}^{\ell}], \quad
\mathbf{V}^{\ell}_k = \mathbb{E}_{\mathbf{X} \sim P_k}[(\mathbf{X}^{\ell} - \overline{\mathbf{X}}^{\ell}_k)^2]
$}
\end{equation}
which capture domain-specific activation patterns shaped by linguistic or semantic properties. A single domain calibration dataset samples only from $P_k(\mathbf{X})$, yielding biased importance metrics that may degrade out-of-domain performance. To mitigate this, we construct a calibration dataset across diverse domains including natural language, source code and mathematical reasoning , ensuring broad coverage of linguistic and logical patterns. The combined calibration distribution is modeled as follows:  
\begin{equation}
\resizebox{0.43\textwidth}{!}{$
P_{\text{calib}}(\mathbf{X}) = \sum_{k=1}^K \alpha_k P_k(\mathbf{X}), 
\alpha_k \geq 0, \ \sum_{k=1}^K \alpha_k = 1
$}
\end{equation}
where $\alpha_k$ reflects each domain’s relative importance. The overall statistics for pruning at layer $\ell$ are defined as follows:  
\begin{equation}
\overline{\mathbf{X}}^{\ell} = \sum_{k=1}^K \alpha_k \overline{\mathbf{X}}^{\ell}_k, \quad
\mathbf{V}^{\ell} = \sum_{k=1}^K \alpha_k \mathbf{V}^{\ell}_k
\end{equation}
providing more representative importance estimates. Calibrating with this domain-diverse dataset enables the pruning algorithm to capture heterogeneous activation behaviors across linguistic and reasoning tasks, yielding more robust and generalizable pruning decisions for large language models.

\subsection{Iterative Calibration Strategy}
During pruning, removing certain channels \( c_k \) in layer \( \ell_i \) inevitably alters the activation statistics of downstream channels \( c_t \) in layers \( \ell_j \) with \( j > i \). Specifically, the baseline activation and variance are defined as follows:  
\begin{equation}
b_t^{(j)} = \mathbb{E}[X_{c_t}^{(\ell_j)}], \quad v_t^{(j)} = \mathrm{Var}[X_{c_t}^{(\ell_j)}]
\end{equation}
Single step calibration methods, such as FLAP, estimate these statistics only once before pruning. For instance, a channel \( c_k \) in \( \ell_i \) may be pruned for low variance \( v_k^{(i)} \), while a channel \( c_t \) in \( \ell_j \) is retained for high variance \( v_t^{(j)} \). However, pruning \( c_k \) and compensating it with a fixed bias replaces its activations with constants, shifting downstream distributions. Consequently, the variance of \( c_t \) may drop sharply as follows:  
\begin{equation}
v_t^{(j)} \rightarrow v_t^{(j)'} \ll v_t^{(j)}
\end{equation}
potentially making \( c_t \) redundant. This reveals a limitation of single-pass calibration: pruning decisions ignore cascading effects from earlier layers. If the pruning mask at step \( s \) is \( M^{(s)} \), then the variance can be expressed as follows:  
\begin{equation}
v_t^{(j, s)} = \mathrm{Var}\big[X_{c_t}^{(\ell_j)} \mid M^{(1)}, \ldots, M^{(s-1)} \big]
\end{equation}
showing that channel variances depend on all prior pruning steps, while single-step methods assume \( s=1 \).

To address this, we introduce an iterative calibration strategy that updates channel importance after each pruning step. At iteration \( s \), recalibrated statistics are computed as follows:  
\begin{equation}
b_t^{(j, s)} = \mathbb{E}[X_{c_t}^{(\ell_j)} \mid M^{(1)}, \ldots, M^{(s-1)}]
\end{equation}
\begin{equation}
v_t^{(j, s)} = \mathrm{Var}[X_{c_t}^{(\ell_j)} \mid M^{(1)}, \ldots, M^{(s-1)}]
\end{equation}
and pruning decisions are based on these refined estimates, allowing dynamically updated importance evaluation. The process continues until a target pruning ratio or convergence criterion is reached. By modeling cascading dependencies, this strategy yields more accurate importance estimation, better global optimization of pruning masks, and improved post-pruning accuracy. Its iterative nature also enables gradual adaptation, reducing reconstruction errors compared with one-shot pruning.

Overall, the iterative calibration can be formulated as minimizing reconstruction error over pruning masks \( M \) as follows:  
\begin{equation}
\min_{M} \mathbb{E}_{\mathbf{X} \sim P_{\text{calib}}} \Big[ \| Y - \widehat{Y}(M; \mathbf{X}) \|^2 \Big]
\end{equation}
where \( Y \) and \( \widehat{Y} \) denote the outputs of the original and pruned models, respectively, and \( M \) is iteratively updated using refined activation statistics.

\section{Experiments}

\subsection{Experimental Setup}
\paragraph{Models and Datasets.}
To assess the effectiveness of our proposed method, we perform experiments on the Qwen2.5 model family, encompassing Qwen2.5-7B, Qwen2.5-14B, and Qwen2.5-32B variants \cite{yang2024qwen2}. We evaluate zero-shot performance on six widely-used commonsense reasoning benchmarks: ARC-Challenge \cite{clark2018ARC}, ARC-Easy \cite{clark2018ARC}, HellaSwag \cite{zellers2019hellaswag}, OpenBookQA (OBQA) \cite{mihaylov2018can}, PIQA \cite{bisk2020piqa}, and Winogrande \cite{ai2:winogrande}.
\paragraph{Baselines.}
We benchmark our approach against two representative structured pruning methods: Wanda-sp \cite{sun2023simple} and FLAP \cite{an2024flap}. It is worth noting that Wanda-sp is an extension of the original Wanda method tailored for structured pruning. 
\paragraph{Implementation Details.}
Our code is implemented using the PyTorch \cite{paszke2019pytorch}  framework and Transformers \cite{wolf2020transformers} libraries, with all experiments conducted on four NVIDIA A100 GPUs. For a fair and comprehensive comparison, all methods are evaluated under two pruning ratios: 25\% and 50\%. All evaluations are conducted using the LM-Harness \cite{eval-harness}.

\begin{table*}[t]
    \centering
    \resizebox{\textwidth}{!}{%
    \begin{tabular}{l|c|c|c|c|c|c|c|c}
        \toprule
        \textbf{Method} & \textbf{Pruning Ratio} & \textbf{ARC-c} & \textbf{ARC-e}  & \textbf{HellaSwag} & \textbf{OBQA} & \textbf{PIQA} & \textbf{Winogrande} & \textbf{Average} \\
        \midrule
        Qwen2.5-14B & 0\%  & 55.8 & 82.49 & 63.38 & 34.4 & 81.12 & 75.3 & 65.42 \\ 
        \midrule 
        Wanda-sp(w\_mix) & \multirow{3}{*}{25\%} & 37.12 & 63.59 & \textbf{46.89} & \textbf{25.0} & \textbf{75.14} & 58.25 & 51.0 \\
        FLAP(w\_mix) &  & 39.51 & 68.39 & 47.42 & 23.8 & 74.86 & 64.72 & 53.12 \\ 
        Ours(w\_mix) &  & \textbf{39.76} & \textbf{68.77} & 46.85 & 24.6 & 74.97 & \textbf{68.67} & \textbf{53.94} \\ 
        \midrule 
        Wanda-sp(w\_mix) & \multirow{3}{*}{50\%} & 21.5 & 27.23 & 25.73 & 14.6 & 54.08 & 49.41 & 32.09 \\ 
        FLAP(w\_mix) &  & 20.99 & 26.22 & 26.26 & 11.4 & 56.09 & 49.49 & 31.74 \\ 
        Ours(w\_mix) &  & \textbf{21.42} & \textbf{39.52} & \textbf{30.49} & \textbf{16.4} & \textbf{62.62} & \textbf{53.67} & \textbf{37.35} \\ 
        \bottomrule
    \end{tabular}
    } 
    \caption{Zero-shot performance of the compressed Qwen2.5-14B. Bold results highlight the best performance.}
    \label{tab:Zero-shot-Qwen2.5-14B}
\end{table*}

\begin{table*}[!htbp]
    \centering
    \resizebox{\textwidth}{!}{%
    \begin{tabular}{l|c|c|c|c|c|c|c|c c}
        \toprule
        \textbf{Method} & \textbf{Pruning Ratio} & \textbf{ARC-c} & \textbf{ARC-e}  & \textbf{HellaSwag} & \textbf{OBQA} & \textbf{PIQA} & \textbf{Winogrande} & \textbf{Average} \\
        \midrule
        Qwen2.5-32 B & 0\%  & 53.41 & 80.51 & 64.91 & 34.2 & 81.88 & 75.3 & 65.04 \\ 
        \midrule 
        Wanda-sp(w\_mix) & \multirow{3}{*}{25\%}           & 42.24 & 70.24 & 52.4 & 27.4 & 76.66 & 61.64 & 55.1 \\ 
        FLAP(w\_mix)  &                                    & 42.24 & 72.85 & 55.02 & 28.6 & 78.02 & 72.53 & 58.21 \\ 
        Ours(w\_mix)  &                    & \textbf{46.67} & \textbf{75.8} & \textbf{57.0} & \textbf{29.6} & \textbf{78.45} & \textbf{72.85} & \textbf{60.06} \\ 
        \midrule 
        Wanda-sp(w\_mix) & \multirow{3}{*}{50\%}           & 24.23 & 32.37 & 27.08 & 15.6 & 57.07 & 50.99 & 34.56 \\ 
        FLAP(w\_mix)  &                                    & 22.7 & 36.36 & 29.43 & 15.6 & 64.36 & 51.07 & 36.59 \\ 
        Ours(w\_mix)  &                    & \textbf{30.72} & \textbf{57.28} & \textbf{39.44} & \textbf{20.2} & \textbf{70.84} & \textbf{61.4} & \textbf{46.65} \\ 
        \bottomrule
    \end{tabular}
    }
    \caption{Zero-shot performance of the compressed Qwen2.5-32B. Bold results highlight the best performance.}
    \label{tab:Zero-shot-Qwen2.5-32B}
\end{table*}

\subsection{Main Results}

As shown in Tables~\ref{tab:Zero-shot-Qwen2.5-14B}  and ~\ref{tab:Zero-shot-Qwen2.5-32B}, our method consistently surpasses existing structured pruning approaches across model scales and compression ratios. The performance gap over FLAP widens with larger models and higher pruning rates, highlighting the scalability and robustness of our approach. Specifically, on Qwen2.5-14B, the gain reaches 6\% at 50\% pruning; and on Qwen2.5-32B, it achieves 1.85\% and 10.06\% improvements at 25\% and 50\%, respectively. These results demonstrate that our iterative calibration effectively preserves task-relevant information and reasoning ability under aggressive compression.

\subsection{Robustness to Calibration Samples}

\begin{figure}[!htbp]
\centering
\begin{subfigure}[b]{0.48\linewidth}
    \centering
    \includegraphics[width=\linewidth]{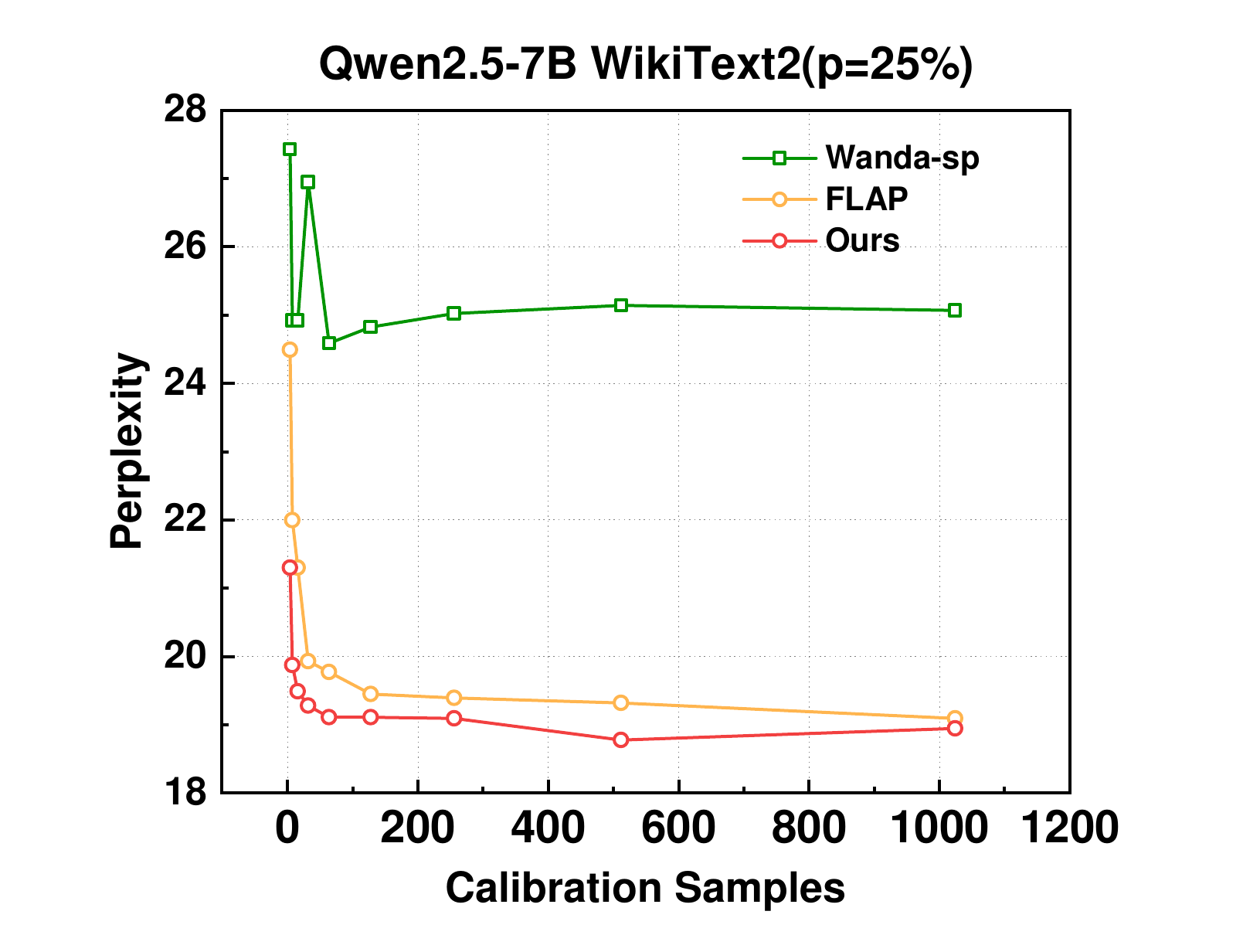}
    \caption{Pruning ratio = 25\% nsamples ablation study}
    \label{fig:nsamples-ablation-0.25}
\end{subfigure}
\hfill
\begin{subfigure}[b]{0.48\linewidth}
    \centering
    \includegraphics[width=\linewidth]{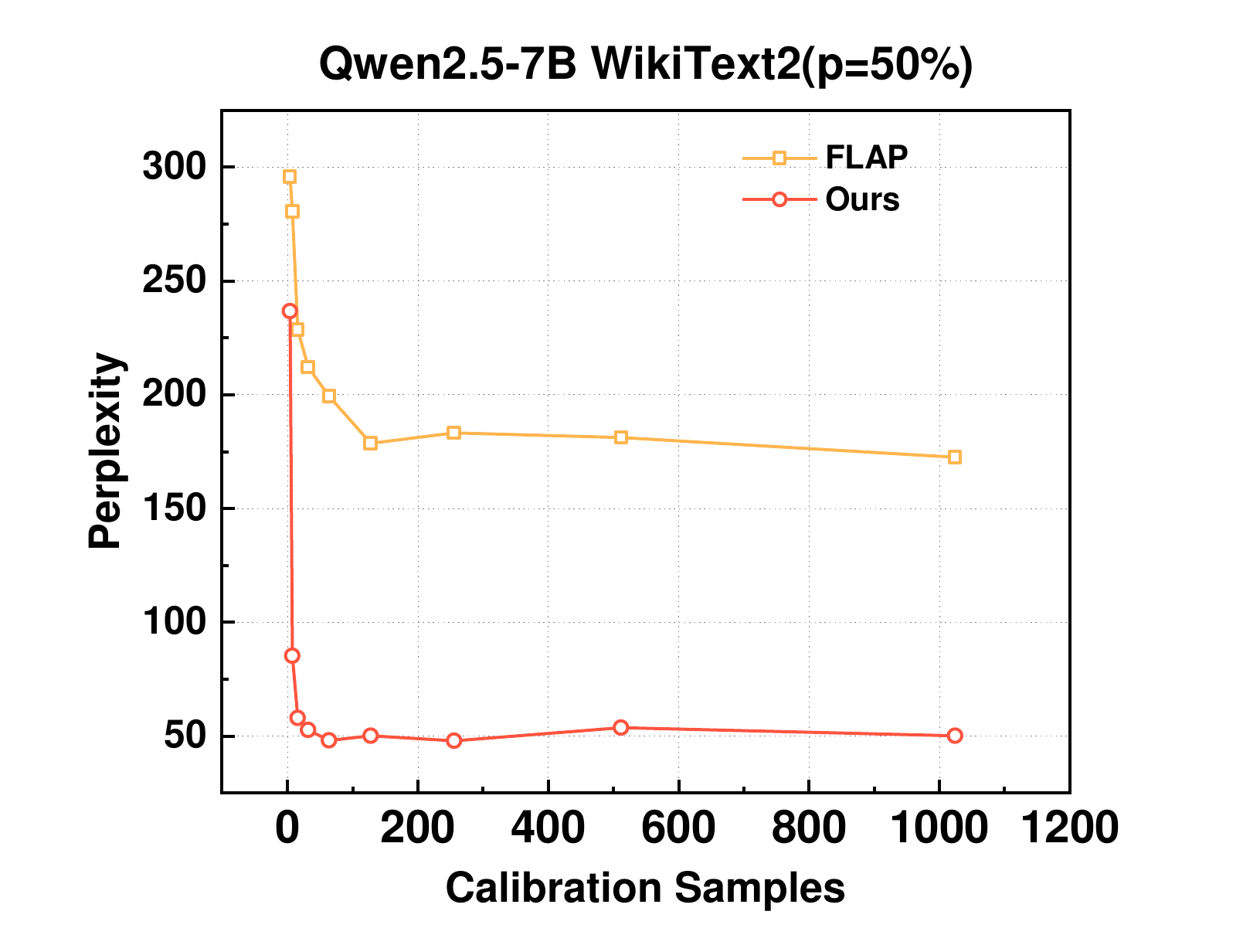}
    \caption{Pruning ratio = 50\% nsamples ablation study}
    \label{fig:nsamples-ablation-0.5}
\end{subfigure}
\caption{Ablation study of nsamples on Qwen2.5-7B under different pruning ratios.}
\label{fig:nsamples-ablation}
\end{figure}

We assess the robustness of our method to the number of calibration samples on Qwen2.5-7B under $25\%$ and $50\%$ pruning using WikiText2. As shown in Figure~\ref{fig:nsamples-ablation-0.25} and Figure~\ref{fig:nsamples-ablation-0.5}, both FLAP and our method benefit from more calibration samples, as reflected in lower perplexity (PPL). Our method consistently outperforms FLAP, with the gap widening at higher pruning ratios. Notably, it achieves PPL $\approx 52$ with only 32 samples and stabilizes near 50 with 128 or more, while FLAP remains above 170 at $50\%$ pruning. These results show that our method better preserves model quality under high sparsity and is more robust to limited calibration data.

\subsection{Different Pruning Ratios}
\begin{figure}[!htbp]
\centering
\begin{subfigure}[b]{0.45\linewidth}
    \centering
    \includegraphics[width=\linewidth]{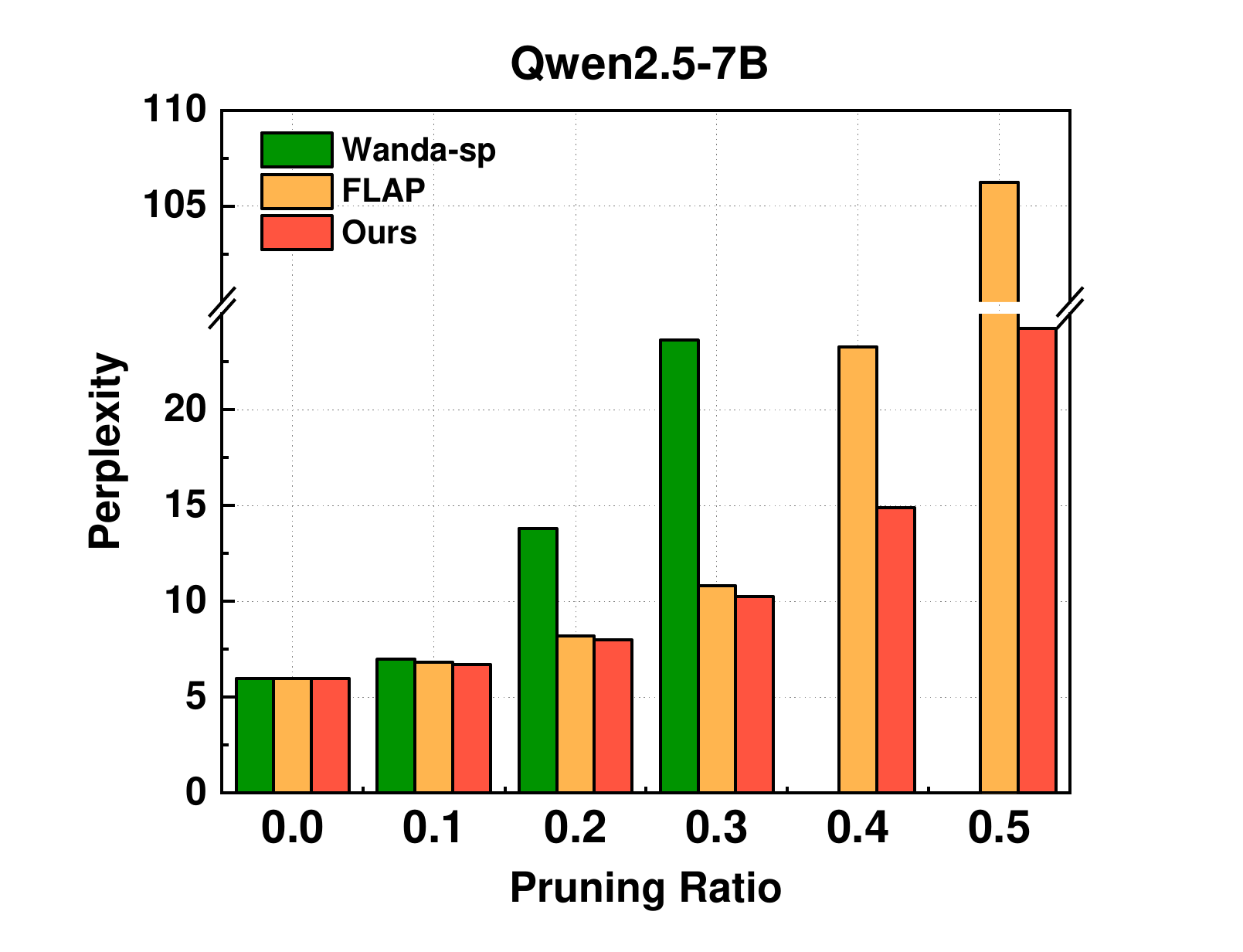}
    \caption{Qwen2.5-7B ratios ablation study}
    \label{fig:ratios-Qwen2.5-7B}
\end{subfigure}
\begin{subfigure}[b]{0.45\linewidth}
    \centering
    \includegraphics[width=\linewidth]{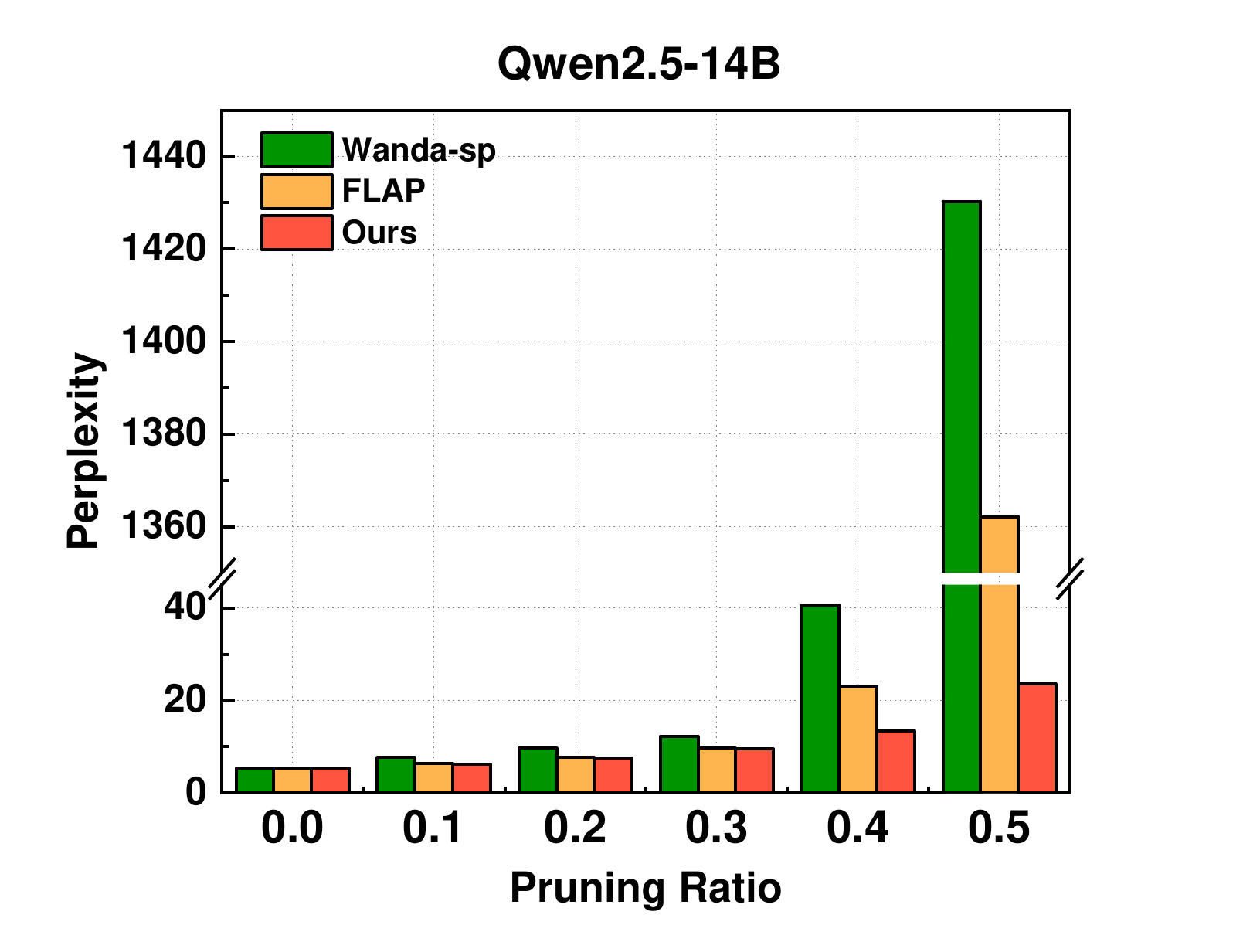}
    \caption{Qwen2.5-14B ratios ablation study}
    \label{fig:ratios-Qwen2.5-14B}
\end{subfigure}

\caption{Ablation studies on pruning ratios for Qwen2.5 models.}
\label{fig:ratios-ablation}
\end{figure}

We evaluate the robustness of our method across pruning ratios on Qwen2.5-7B and Qwen2.5-14B, comparing with Wanda-sp and FLAP. As shown in Figure~\ref{fig:ratios-Qwen2.5-7B} and Figure~\ref{fig:ratios-Qwen2.5-14B}, our method consistently outperforms both baselines, with the advantage increasing as pruning becomes more aggressive. On Qwen2.5-7B, at $50\%$ pruning, Wanda-sp collapses (PPL $>6800$) and FLAP degrades severely (PPL $>106$), while our method maintains a low PPL of 24.2. A similar pattern appears on Qwen2.5-14B, where Wanda-sp and FLAP reach PPLs of 1430 and 1362, respectively, whereas our method achieves only 23.7. These results confirm that our iterative compensation strategy enables stable, high-quality performance even under extreme sparsity.

\begin{table*}[!t]
    \centering
    \resizebox{\textwidth}{!}{%
    \begin{tabular}{l|c|c|c|c|c|c|c|c c}
        \toprule
        \textbf{Method} & \textbf{Pruning Ratio} & \textbf{ARC-c} & \textbf{ARC-e}  & \textbf{HellaSwag} & \textbf{OBQA} & \textbf{PIQA} & \textbf{Winogrande} & \textbf{Average} \\
        \midrule
        Qwen2.5-14B & 0\%  & 55.8 & 82.49 & 63.38 & 34.4 & 81.12 & 75.3 & 65.42 \\ 
        \midrule 
        Ours  & \multirow{2}{*}{25\%}           & \textbf{41.64} & \textbf{70.5} & 44.73 & \textbf{28.0} & 71.16 & 67.72 & \textbf{53.96} \\ 
        Ours(w\_mix)  &                    & 39.76 & 68.77 & \textbf{46.85} & 24.6 & \textbf{74.97} & \textbf{68.67} & 53.94 \\ 
        \midrule 
        Ours  & \multirow{2}{*}{50\%}           & 20.48 & 39.18 & 29.14 & \textbf{16.8} & 58.92 & 50.91 & 35.9 \\ 
        Ours(w\_mix)  &                    & \textbf{21.42} & \textbf{39.52} & \textbf{30.49} & 16.4 & \textbf{62.62} & \textbf{53.67} & \textbf{37.35} \\ 
        \bottomrule
    \end{tabular}
    }
     \caption{Performance Comparsion of the compressed Qwen2.5-14B with and without  multi-domain hybrid calibration set. Bold results highlight the best performance.}
     \label{tab:mix-ablation-Qwen2.5-14B}
\end{table*}

\begin{figure*}[!htbp]
\centering
\begin{subfigure}[b]{0.3\linewidth}
    \centering
    \includegraphics[width=\linewidth]{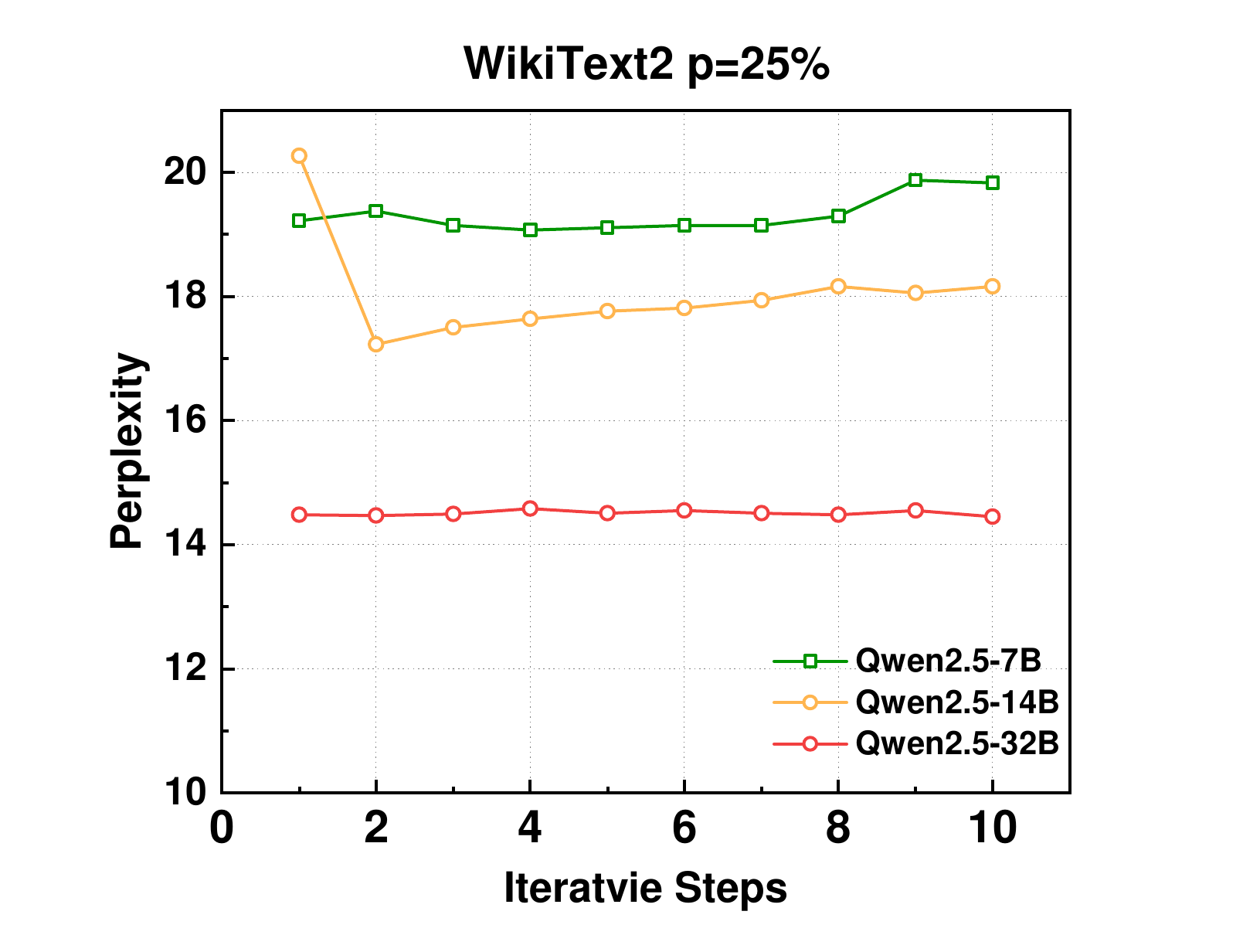}
    \caption{Pruning ratio = 25\%}
    \label{fig:iter-steps-25}
\end{subfigure}
\hfill
\begin{subfigure}[b]{0.3\linewidth}
    \centering
    \includegraphics[width=\linewidth]{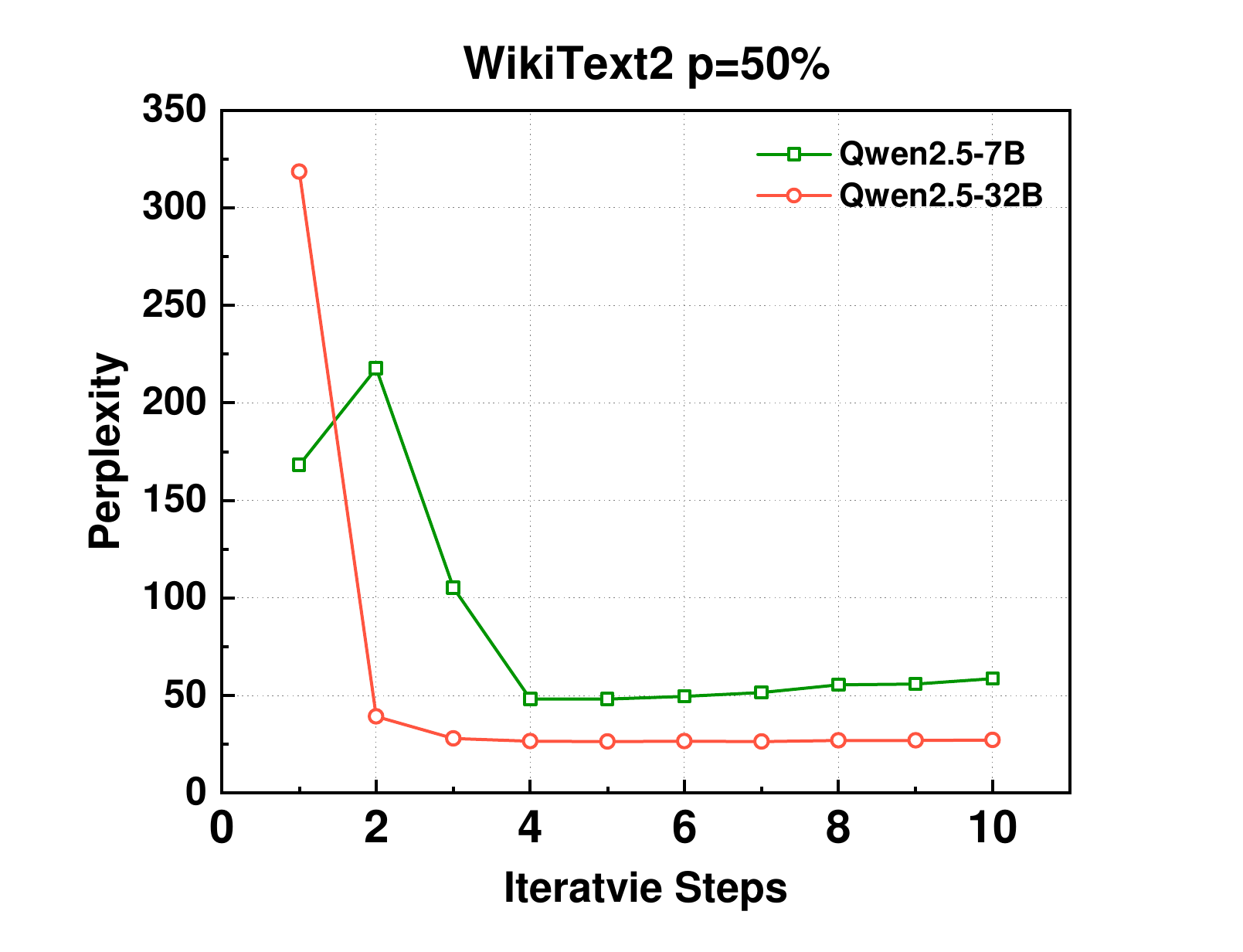}
    \caption{Pruning ratio = 50\%}
    \label{fig:iter-steps-50}
\end{subfigure}
\hfill
\begin{subfigure}[b]{0.3\linewidth}
    \centering
    \includegraphics[width=\linewidth]{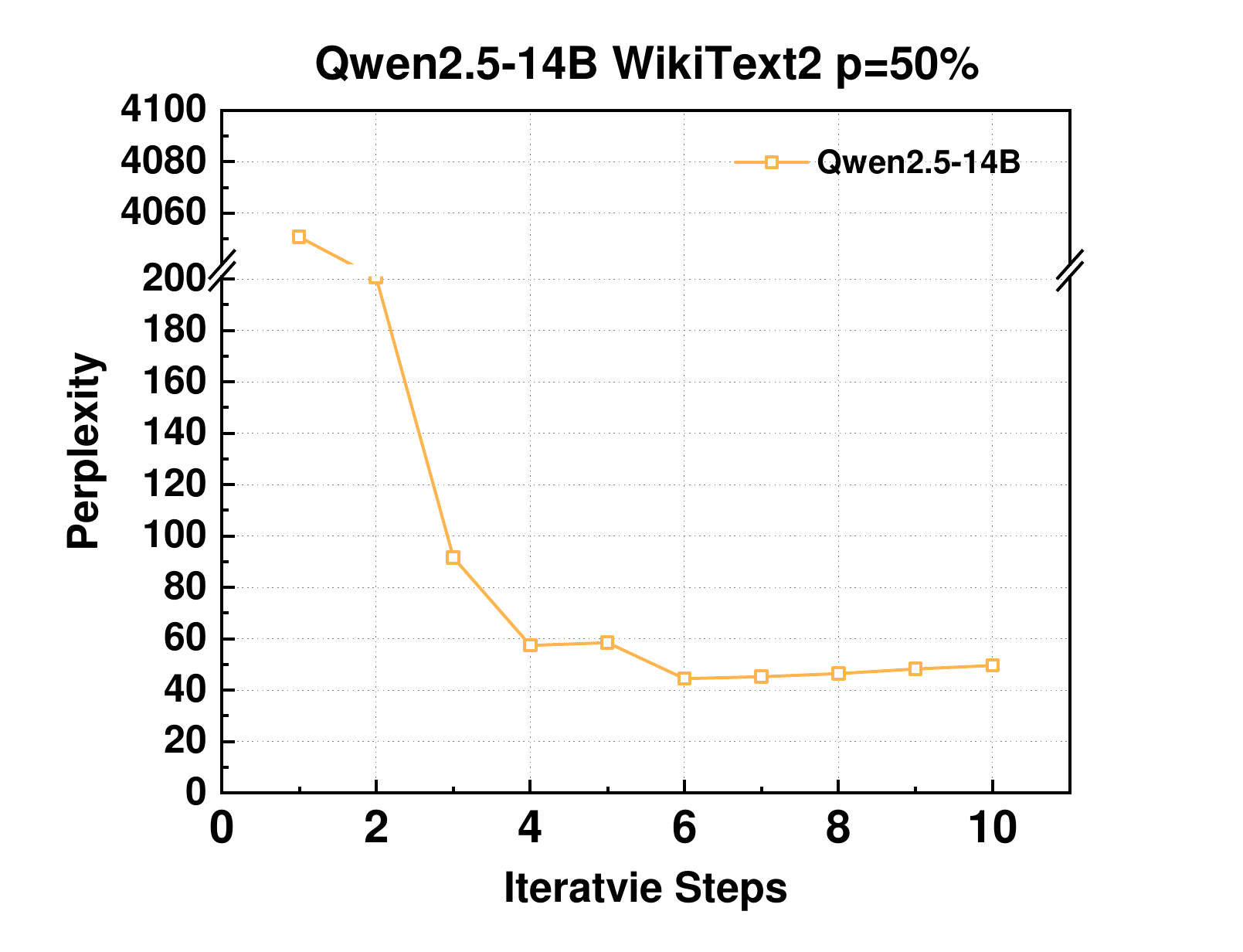}
    \caption{Qwen2.5-14B}
    \label{fig:iter-steps-14b}
\end{subfigure}
\caption{Ablation studies on iterative pruning steps across different pruning ratios and models.}
\label{fig:iter-steps-ablation}
\end{figure*}
\subsection{Ablation Study}
To comprehensively analyze the individual contribution of each component in our proposed framework, we conducted a series of ablation studies. These experiments specifically investigate the effectiveness of incorporating a multi-domain hybrid calibration set, as well as systematically assess the impact of the iterative pruning strategy.

\paragraph{Multi-domain Hybrid Calibration Set. }

Activation statistics (e.g., channel-wise mean and variance) vary across data domains, affecting pruning accuracy. To address this, we introduce a multi-domain hybrid calibration set to capture broader activation variations. We evaluate this design on Qwen2.5-14B under $25\%$ and $50\%$ pruning, comparing single-domain calibration with our hybrid approach. As shown in Tables~\ref{tab:mix-ablation-Qwen2.5-14B}, the hybrid setting consistently outperforms the single-domain variant, achieving higher zero-shot accuracy on average. These results confirm that multi-domain calibration provides more robust channel importance estimation and improves structured pruning performance.

\paragraph{Iterative Pruning. }
We study the effect of iterative pruning steps on model quality using Qwen2.5-7B, Qwen2.5-14B, and Qwen2.5-32B with WikiText2 calibration under $25\%$ and $50\%$ pruning. As shown in Figure~\ref{fig:iter-steps-ablation}, model perplexity remains stable across step counts at $25\%$ pruning, indicating low sensitivity in this regime. In contrast, at $50\%$ pruning, iterative pruning significantly improves performance: perplexity decreases with more steps, especially within the first three to four iterations. For instance, on Qwen2.5-14B, single-shot pruning causes severe degradation , while six iterative steps reduce it to about 44. These results clearly show that gradual, multi-step pruning is crucial for maintaining quality under high sparsity, and that four to six iterations are typically sufficient to achieve most of the gains, consistently across all evaluated datasets.

\section{Conclusion}
In this work, we introduce a novel structured pruning framework that synergistically integrates a multi-domain hybrid calibration set with an iterative, progressive pruning strategy. This design facilitates more precise identification of redundant channels while maintaining model performance across a wide spectrum of tasks. Comprehensive evaluations on multiple state-of-the-art large language models demonstrate that our approach consistently surpasses existing baselines, achieving substantial compression with minimal degradation in accuracy. These findings underscore the critical role of diverse calibration data and gradual pruning schedules in enabling efficient model compression.

\section*{Limitations}

In this work, we conduct extensive experiments to evaluate the effectiveness of our pruning method. The results demonstrate that our approach achieves competitive performance compared to the baselines. However, due to computational constraints, we have not yet been able to evaluate it on larger scale models, such as those with 70 billion parameters. Exploring the scalability of our method to such large models constitutes an important direction for future work.

\bibliography{custom}

@article{qin2023,
  author = {Qin, A. and others},
  year = {2023},
  title = {Advances in state-of-the-art natural language processing},
  journal = {Journal of NLP Research},
}

@article{zhu2023,
  author = {Zhu, B. and others},
  year = {2023},
  title = {Large Language Models: Progress and Applications},
  journal = {Advances in NLP},
}

@article{li2023,
  author = {Li, C. and others},
  year = {2023},
  title = {Fine-Tuning Techniques for Efficient Model Adaptation},
  journal = {AI Research Journal},
}

@article{zhang2023,
  author = {Zhang, D. and others},
  year = {2023},
  title = {Parameter-Efficient Fine-Tuning Methods for LLMs},
  journal = {Journal of Machine Learning Research},
}

@article{huang2023,
  author = {Huang, E. and others},
  year = {2023},
  title = {Evaluating Large Language Models in Complex Scenarios},
  journal = {Journal of Computational Linguistics},
}

@article{wang2023,
  author = {Wang, F. and others},
  year = {2023},
  title = {Practical Applications of LLMs in Specialized Domains},
  journal = {Specialized AI Applications},
}

@article{ding2022,
  author = {Ding, G. and others},
  year = {2022},
  title = {Efficient Fine-Tuning for Resource-Constrained Systems},
  journal = {Proceedings of the Machine Learning Conference},
}

@article{paszke2019pytorch,
  title={Pytorch: An imperative style, high-performance deep learning library},
  author={Paszke, Adam and Gross, Sam and Massa, Francisco and Lerer, Adam and Bradbury, James and Chanan, Gregory and Killeen, Trevor and Lin, Zeming and Gimelshein, Natalia and Antiga, Luca and others},
  journal={Advances in neural information processing systems},
  volume={32},
  year={2019}
}

@article{wolf2020transformers,
  title={Transformers: State-of-the-Art Natural Language Processing},
  author={Wolf, Thomas},
  journal={arXiv preprint arXiv:1910.03771},
  year={2020}
}

@article{ma2023llm,
  title={Llm-pruner: On the structural pruning of large language models},
  author={Ma, Xinyin and Fang, Gongfan and Wang, Xinchao},
  journal={Advances in neural information processing systems},
  volume={36},
  pages={21702--21720},
  year={2023}
}

@article{ashkboos2024slicegpt,
  title={Slicegpt: Compress large language models by deleting rows and columns},
  author={Ashkboos, Saleh and Croci, Maximilian L and Nascimento, Marcelo Gennari do and Hoefler, Torsten and Hensman, James},
  journal={arXiv preprint arXiv:2401.15024},
  year={2024}
}

@article{li2023communication,
  title={A Communication-Efficient, Privacy-Preserving Federated Learning Algorithm Based on Two-Stage Gradient Pruning and Differentiated Differential Privacy},
  author={Li, Yong and Du, Wei and Han, Liquan and Zhang, Zhenjian and Liu, Tongtong},
  journal={Sensors},
  volume={23},
  number={23},
  pages={9305},
  year={2023},
  publisher={MDPI}
}

@article{han2015deep,
  title={Deep compression: Compressing deep neural networks with pruning, trained quantization and huffman coding},
  author={Han, Song and Mao, Huizi and Dally, William J},
  journal={arXiv preprint arXiv:1510.00149},
  year={2015}
}

@article{ai2:winogrande,
  title={Winogrande: An adversarial winograd schema challenge at scale},
  author={Sakaguchi, Keisuke and Bras, Ronan Le and Bhagavatula, Chandra and Choi, Yejin},
  journal={Communications of the ACM},
  volume={64},
  number={9},
  pages={99--106},
  year={2021},
  publisher={ACM New York, NY, USA}
}

@article{mihaylov2018can,
  title={Can a suit of armor conduct electricity? a new dataset for open book question answering},
  author={Mihaylov, Todor and Clark, Peter and Khot, Tushar and Sabharwal, Ashish},
  journal={arXiv preprint arXiv:1809.02789},
  year={2018}
}

@article{sun2023simple,
  title={A simple and effective pruning approach for large language models},
  author={Sun, Mingjie and Liu, Zhuang and Bair, Anna and Kolter, J Zico},
  journal={arXiv preprint arXiv:2306.11695},
  year={2023}
}

@article{yang2021knowledge,
  title={Knowledge distillation: A survey},
  author={Yang, Zhen and Zhang, Zilun and Wang, Sheng and Li, Jie and Zhang, Meishan and Liu, Zhiyuan and Sun, Maosong},
  journal={arXiv preprint arXiv:2106.05860},
  year={2021}
}

@article{zhou2021smoothquant,
  title={SmoothQuant: Accurate and Efficient Post-Training Quantization for Large Language Models},
  author={Zhou, Yuxiao and Wang, Zhen and Li, Yujun and Wang, Sheng and Liu, Zhiyuan and Sun, Maosong},
  journal={arXiv preprint arXiv:2302.06557},
  year={2023}
}

@article{cai2023gptq,
  title={GPTQ: Accurate Post-Training Quantization for Generative Pre-trained Transformers},
  author={Cai, Yuchen and Wang, Zhen and Li, Yujun and Wang, Sheng and Liu, Zhiyuan and Sun, Maosong},
  journal={arXiv preprint arXiv:2302.06557},
  year={2023}
}

@article{zhou2024framequant,
  title={FrameQuant: Flexible Low-Bit Quantization for Transformers},
  author={Zhou, Yuxiao and Wang, Zhen and Li, Yujun and Wang, Sheng and Liu, Zhiyuan and Sun, Maosong},
  journal={arXiv preprint arXiv:2402.06557},
  year={2024}
}

@misc{2024CDL,
  title={Continual Distillation Learning: Knowledge Distillation in Prompt-based Continual Learning},
  author={Qifan Zhang and Yunhui Guo and Yu Xiang},
  year={2024},
  eprint={2407.13911},
  archivePrefix={arXiv},
  primaryClass={cs.CV}
}

@inproceedings{yang2022comparative,
  title={Comparative Analysis of Structured Pruning and Unstructured Pruning},
  author={Yang, Zhengwu and Zhang, Han},
  booktitle={Frontier Computing},
  pages={112},
  year={2022},
  organization={Springer},
  doi={10.1007/978-981-16-8052-6_112}
}

@inproceedings{liao2023unstructured,
  title={Can Unstructured Pruning Reduce the Depth in Deep Neural Networks?},
  author={Liao, Sheng and others},
  booktitle={Proceedings of the IEEE/CVF Conference on Computer Vision and Pattern Recognition Workshops},
  year={2023}
}

@article{unstructured_pruning_speech,
  title={Unstructured Pruning and Low Rank Factorisation of Self-Supervised Pre-trained Speech Models},
  author={Anonymous},
  journal={IEEE Transactions on Audio, Speech, and Language Processing},
  volume={},
  number={},
  pages={1046-1058},
  year={2024},
  publisher={IEEE}
}

@article{yang2024qwen2,
  title={Qwen2. 5 technical report},
  author={Yang, An and Yang, Baosong and Zhang, Beichen and Hui, Binyuan and Zheng, Bo and Yu, Bowen and Li, Chengyuan and Liu, Dayiheng and Huang, Fei and Wei, Haoran and others},
  journal={arXiv preprint arXiv:2412.15115},
  year={2024}
}

@inproceedings{bisk2020piqa,
  title={Piqa: Reasoning about physical commonsense in natural language},
  author={Bisk, Yonatan and Zellers, Rowan and Gao, Jianfeng and Choi, Yejin and others},
  booktitle={Proceedings of the AAAI conference on artificial intelligence},
  volume={34},
  pages={7432--7439},
  year={2020}
}

@article{zellers2019hellaswag,
  title={Hellaswag: Can a machine really finish your sentence?},
  author={Zellers, Rowan and Holtzman, Ari and Bisk, Yonatan and Farhadi, Ali and Choi, Yejin},
  journal={arXiv preprint arXiv:1905.07830},
  year={2019}
}

@article{clark2018ARC,
  title={Think you have solved question answering? try arc, the ai2 reasoning challenge},
  author={Clark, Peter and Cowhey, Isaac and Etzioni, Oren and Khot, Tushar and Sabharwal, Ashish and Schoenick, Carissa and Tafjord, Oyvind},
  journal={arXiv preprint arXiv:1803.05457},
  year={2018}
}

@inproceedings{an2024flap,
  title={Fluctuation-based adaptive structured pruning for large language models},
  author={An, Yongqi and Zhao, Xu and Yu, Tao and Tang, Ming and Wang, Jinqiao},
  booktitle={Proceedings of the AAAI Conference on Artificial Intelligence},
  volume={38},
  pages={10865--10873},
  year={2024}
}

@misc{eval-harness,
  author       = {Gao, Leo and Tow, Jonathan and Abbasi, Baber and Biderman, Stella and Black, Sid and DiPofi, Anthony and Foster, Charles and Golding, Laurence and Hsu, Jeffrey and Le Noac'h, Alain and Li, Haonan and McDonell, Kyle and Muennighoff, Niklas and Ociepa, Chris and Phang, Jason and Reynolds, Laria and Schoelkopf, Hailey and Skowron, Aviya and Sutawika, Lintang and Tang, Eric and Thite, Anish and Wang, Ben and Wang, Kevin and Zou, Andy},
  title        = {The Language Model Evaluation Harness},
  month        = 07,
  year         = 2024,
  publisher    = {Zenodo},
  version      = {v0.4.3},
  doi          = {10.5281/zenodo.12608602},
  url          = {https://zenodo.org/records/12608602}
}

@article{ma2025cad,
  title={CAD-VAE: Leveraging Correlation-Aware Latents for Comprehensive Fair Disentanglement},
  author={Ma, Chenrui and Zhao, Rongchang and Xiao, Xi and Xie, Hongyang and Wang, Tianyang and Wang, Xiao and Zhang, Hao and Shen, Yanning},
  journal={arXiv preprint arXiv:2503.07938},
  year={2025}
}

@article{wu2025sugar,
  title={Sugar-coated poison: Benign generation unlocks llm jailbreaking},
  author={Wu, Yu-Hang and Xiong, Yu-Jie and Zhang, Hao and Zhang, Jia-Chen and Zhou, Zheng},
  journal={EMNLP 2025 Findings},
  year={2025}
}

@inproceedings{qi2025mediaug,
  title={Mediaug: Exploring visual augmentation in medical imaging},
  author={Qi, Xuyin and Zhang, Zeyu and Gang, Canxuan and Zhang, Hao and Zhang, Lei and Zhang, Zhiwei and Zhao, Yang},
  booktitle={Annual Conference on Medical Image Understanding and Analysis},
  pages={218--232},
  year={2025},
  organization={Springer}
}

@article{qi2025medconv,
  title={Medconv: Convolutions beat transformers on long-tailed bone density prediction},
  author={Qi, Xuyin and Zhang, Zeyu and Zheng, Huazhan and Chen, Mingxi and Kutaiba, Numan and Lim, Ruth and Chiang, Cherie and Tham, Zi En and Ren, Xuan and Zhang, Wenxin and others},
  journal={IJCNN2025},
  year={2025}
}

@article{luo2025pathohr,
  title={Pathohr: Breast cancer survival prediction on high-resolution pathological images},
  author={Luo, Yang and Wang, Shiru and Liu, Jun and Xiao, Jiaxuan and Xue, Rundong and Zhang, Zeyu and Zhang, Hao and Lu, Yu and Zhao, Yang and Xie, Yutong},
  journal={arXiv preprint arXiv:2503.17970},
  year={2025}
}

@article{zhang2025can,
  title={Can Representation Gaps Be the Key to Enhancing Robustness in Graph-Text Alignment?},
  author={Zhang, Heng and Zhang, Tianyi and Shi, Yuling and Gu, Xiaodong and Shen, Yaomin and Zhang, Zijian and Yuan, Yilei and Zhang, Hao and Huang, Jin},
  journal={arXiv preprint arXiv:2510.12087},
  year={2025}
}

@article{zhang2025asymoe,
  title={AsyMoE: Leveraging Modal Asymmetry for Enhanced Expert Specialization in Large Vision-Language Models},
  author={Zhang, Heng and Hu, Haichuan and Shen, Yaomin and Yu, Weihao and Yuan, Yilei and You, Haochen and Cheng, Guo and Zhang, Zijian and Gan, Lubin and Wei, Huihui and others},
  journal={arXiv preprint arXiv:2509.12715},
  year={2025}
}

@article{zheng2025g2rammar,
  title={G2rammar: Bilingual Grammar Modeling for Enhanced Text-attributed Graph Learning},
  author={Zheng, Heng and You, Haochen and Liu, Zijun and Zhang, Zijian and Gan, Lubin and Zhang, Hao and Huang, Wenjun and Huang, Jin},
  journal={arXiv preprint arXiv:2511.00911},
  year={2025}
}

@article{zheng2025graphgeo,
  title={GraphGeo: Multi-Agent Debate Framework for Visual Geo-localization with Heterogeneous Graph Neural Networks},
  author={Zheng, Heng and Shi, Yuling and Gu, Xiaodong and You, Haochen and Zhang, Zijian and Gan, Lubin and Zhang, Hao and Huang, Wenjun and Huang, Jin},
  journal={arXiv preprint arXiv:2511.00908},
  year={2025}
}

@inproceedings{cong2025hierarchical,
  title={Hierarchical Multi-Scale Feature Fusion Network for Multi-Center Major Depressive Disorder Classification with T1-weighted MRI.},
  author={Cong, Zhaoyang and Wang, Ziyang and Zhang, Hao and Zheng, Guowei and Cao, Keming and Zhao, Lina and Song, Ruipeng and Li, Jianqing and Liu, Chengyu},
  booktitle={Annual International Conference of the IEEE Engineering in Medicine and Biology Society. IEEE Engineering in Medicine and Biology Society. Annual International Conference},
  volume={2025},
  pages={1--4},
  year={2025}
}
\clearpage
\appendix

\section{Comparison Experiments on Qwen2.5-7B}

\begin{table*}[htbp]
    \centering
    \resizebox{\textwidth}{!}{%
    \begin{tabular}{l|c|c|c|c|c|c|c|c c}
        \toprule
        \textbf{Method} & \textbf{Pruning Ratio} & \textbf{ARC-c} & \textbf{ARC-e}  & \textbf{HellaSwag} & \textbf{OBQA} & \textbf{PIQA} & \textbf{Winogrande} & \textbf{Average} \\
        \midrule
        Qwen2.5-32B & 0\%  & 53.41 & 80.51 & 64.91 & 34.2 & 81.88 & 75.3 & 65.04 \\ 
        \midrule 
        Ours  & \multirow{2}{*}{25\%}           & 46.08 & 74.87 & 53.35 & \textbf{30.6} & 75.35 & \textbf{73.32} & 58.93 \\ 
        Ours(w\_mix)  &                    & \textbf{46.67} & \textbf{75.8} & \textbf{57.0} & 29.6 & \textbf{78.45} & 72.85 & \textbf{60.06} \\ 
        \midrule
        Ours  & \multirow{2}{*}{50\%}           & 29.01 & 57.28 & 36.89 & \textbf{23.6} & 65.18 & 58.88 & 45.14 \\ 
        Ours(w\_mix)  &                    & \textbf{30.72} & 57.28 & \textbf{39.44} & 20.2 & \textbf{70.84} & \textbf{61.4} & \textbf{46.65} \\ 
        \bottomrule
    \end{tabular}
    }
    \caption{Performance Comparsion of the compressed Qwen2.5-32B with and without  multi-domain hybrid calibration set. Bold results highlight the best performance.}
    \label{tab:mix-ablation-Qwen2.5-32B}
\end{table*}
We also conducted experiments on Qwen2.5-7B across multiple datasets. As shown in Table~\ref{tab:Zero-shot-Qwen2.5-7B}, our method consistently achieves strong performance, demonstrating the effectiveness and general applicability of our pruning approach.
\begin{table*}[t]
    \centering
    \resizebox{\textwidth}{!}{%
    \begin{tabular}{l|c|c|c|c|c|c|c|c @{} }
        \toprule
        \textbf{Method} & \textbf{Pruning Ratio} & \textbf{ARC-c} & \textbf{ARC-e} & \textbf{HellaSwag} & \textbf{OBQA} & \textbf{PIQA} & \textbf{Winogrande} & \textbf{Average} \\
        \midrule
        Qwen2.5-7 B & 0\%  & 47.61 & 80.47 & 59.95 & 33.8 & 78.56 & 72.85 & 62.21 \\ 
        \midrule 
        Wanda-sp(w\_mix) & \multirow{3}{*}{25\%}           & 33.62 & 63.22 & \textbf{43.45} & 23.8 & \textbf{73.23} & 54.06 & 48.56 \\ 
        FLAP(w\_mix)  &                                    & 32.08 & 62.33 & 41.75 & 21.4 & 72.31 & 59.59 & 48.24 \\ 
        Ours(w\_mix)  &                    & \textbf{34.04} & \textbf{65.45} & 43.12 & \textbf{24.6} & 72.85 & \textbf{60.54} & \textbf{50.1} \\ 
        \midrule 
        Wanda-sp(w\_mix) & \multirow{3}{*}{50\%}           & \textbf{21.67} & 25.59 & 25.64 & \textbf{14.6} & 51.85 & \textbf{51.78} & 31.85 \\ 
        FLAP(w\_mix)  &                                    & 19.37 & 29.97 & 27.17 & 12.2 & 56.09 & 49.01 & 32.3 \\ 
        Our method(w\_mix)  &                    & 18.86 & \textbf{35.4} & \textbf{29.35} & 12.4 & \textbf{60.77} & 50.2 & \textbf{34.49} \\ 
        \bottomrule
    \end{tabular}
    }
    \caption{Zero-shot performance of the compressed Qwen2.5-7B. Bold results highlight the best performance.}
    \label{tab:Zero-shot-Qwen2.5-7B}
\end{table*}

\section{Ablation of Multi-Domain Calibration on Qwen2.5-32B}
We evaluate multi domain calibration on Qwen2.5-32B under $25\%$ and $50\%$ pruning, comparing single-domain calibration with our hybrid approach. As shown in Tables~\ref{tab:mix-ablation-Qwen2.5-32B}, the hybrid setting consistently outperforms the single-domain variant, achieving higher zero-shot accuracy on average. These results confirm that multi-domain calibration provides more robust channel importance estimation and improves structured pruning performance.

\end{document}